\documentclass[letterpaper]{article} 
\usepackage{aaai25}  

\usepackage{times}  
\usepackage{helvet}  
\usepackage{courier}  
\usepackage[hyphens]{url}  
\usepackage{graphicx} 
\usepackage{natbib}  
\usepackage{caption} 
\usepackage{complexity}
\usepackage{bm}         
\usepackage{amssymb}
\usepackage{amsthm}
\usepackage{amsmath}
\usepackage{enumitem}
\usepackage{caption}
\usepackage{subcaption}
\usepackage{centernot}
\usepackage{complexity}
\usepackage{mathrsfs}   
\usepackage{tabularx}   
\usepackage{multirow}   
\usepackage{array}      
\usepackage{booktabs}   
\usepackage{colortbl}   
\usepackage{xargs}      
\usepackage{tikz}
\usepackage{tikz-cd}
\usepackage{url}
\usepackage{makecell}   
\usepackage{float}
\usepackage{etoc}
\usepackage[ruled,noline,linesnumbered,noend]{algorithm2e} 
\usepackage{lipsum}
\usepackage{float}
\usepackage{adjustbox}
\usepackage{placeins}
\usepackage{listings}

\definecolor{codegreen}{rgb}{0,0.6,0}
\definecolor{codegray}{rgb}{0.5,0.5,0.5}
\definecolor{codepurple}{rgb}{0.58,0,0.82}
\definecolor{backcolour}{rgb}{0.95,0.95,0.92}

\lstdefinestyle{mystyle}{
    belowskip=\smallskipamount,
    belowcaptionskip=0mm,
    backgroundcolor=\color{backcolour},   
    commentstyle=\color{codegreen},
    keywordstyle=\color{magenta},
    stringstyle=\color{codepurple},
    basicstyle=\fontencoding{T1}\scriptsize\fontfamily{lmtt}\fontseries{m}\selectfont,
    breakatwhitespace=false,         
    breaklines=true,                 
    captionpos=b,                    
    keepspaces=true,                 
    showspaces=false,                
    showstringspaces=false,
    showtabs=false,   
    framexleftmargin=1mm, 
    xleftmargin=1mm,             
    tabsize=2
}

\usetikzlibrary{calc,positioning}

\newtheorem{theorem}{Theorem}[section]

\theoremstyle{definition}

\theoremstyle{definition}
\newtheorem{definition}[theorem]{Definition}

\SetInd{0.5em}{0em}

\newcolumntype{Y}{>{\raggedleft\arraybackslash}X}
\newcommand{\header}[1]{\rotatebox[origin=l]{90}{\hspace*{-0.22cm} #1}}

\newcommand{\zerocell}[1]{-}

\definecolor{caribbeangreen}{rgb}{0.0, 0.8, 0.6}
\definecolor{brilliantlavender}{rgb}{0.96, 0.73, 1.0}
\definecolor{amethyst}{rgb}{0.6, 0.4, 0.8}
\definecolor{ao(english)}{rgb}{0.0, 0.5, 0.0}
\definecolor{arylideyellow}{rgb}{0.91, 0.84, 0.42}
\definecolor{asparagus}{rgb}{0.53, 0.66, 0.42}
\definecolor{aquamarine}{rgb}{0.5, 1.0, 0.83}
\definecolor{babyblue}{rgb}{0.54, 0.81, 0.94}
\definecolor{fwtchanged}{rgb}{0.3, 0.3, 0.7}
\definecolor{rosewood}{rgb}{0.4, 0.0, 0.04}
\definecolor{oldmauve}{rgb}{0.4, 0.19, 0.28}
\definecolor{myrtle}{rgb}{0.13, 0.26, 0.12}
\definecolor{magenta(dye)}{rgb}{0.79, 0.08, 0.48}

\definecolor{plta}{rgb}{0.12, 0.47, 0.71}
\definecolor{pltb}{rgb}{   1, 0.5, 0.05}
\definecolor{pltc}{rgb}{0.17, 0.63, 0.17}
\definecolor{pltd}{rgb}{0.84, 0.15, 0.16}

\newcommand{\sssection}[1]{\subsubsection{#1}}

\newcommand{\colourdc}{blue}
  
\newcommand{\todocustom}[3]{\todo[linecolor=#2,backgroundcolor=#2!25,bordercolor=#2,size=\tiny,#3]{#1}}  
\newcommandx{\nbdc}[2][1=]{\todocustom{#2}{\colourdc}{#1}}

\def\X{\mathbf{X}}
\def\N{\mathbb{N}}
\def\R{\mathbb{R}}

\def\s{\sigma}

\renewcommand{\phi}{\varphi}

\def\la{\leftarrow}

\newcommand{\abs}[1]{\left| #1 \right|}

\newcommand{\gen}[1]{\left< #1 \right>}

\newcommand{\set}[1]{\left\{ #1 \right\}}
\newcommand{\seta}[1]{\{ #1 \}}

\newcommand{\mseta}[1]{\{ \!\! \{ #1 \} \!\! \}}

\newcommand{\lr}[1]{\left( #1 \right)}

\newcommand{\range}[1]{\left[\!\left[ #1 \right]\!\right]}

\DeclareMathOperator*{\concat}{%
  \mathchoice%
    {\Big\Vert}%
    {\big\Vert}%
    {\Vert}%
    {\Vert}%
}

\newcommand{\objects}{\mathcal{O}}
\newcommand{\predicates}{\Sigma_{p}}
\newcommand{\functions}{\Sigma_{f}}
\newcommand{\schemata}{\mathcal{A}}
\newcommand{\goal}{\mathcal{G}}
\newcommand{\objs}{\mathbf{o}}

\newcommand{\arity}{\mathrm{ar}}

\newcommand{\hash}{\textsc{hash}}
\newcommand{\graphFont}[1]{\mathbf{#1}}
\newcommand{\neighbour}{\graphFont{N}}
\newcommand{\featCat}{\graphFont{F}_{\text{cat}}}
\newcommand{\featCon}{\graphFont{F}_{\text{con}}}
\newcommand{\featCol}{\graphFont{F}_{\text{clr}}}
\newcommand{\featEdge}{\graphFont{L}}
\newcommand{\graph}{\graphFont{G}}
\newcommand{\nodes}{\graphFont{V}}
\newcommand{\edges}{\graphFont{E}}

\newcommand{\nodeCat}{\Sigma_{\text{V}}}
\newcommand{\edgeCat}{\Sigma_{\text{E}}}

\newcommand{\colours}{\graphFont{C}}
\newcommand{\countCol}{\textsc{count}}

\newcommand{\Mst}{\graphFont{M}}

\newcommand{\nilg}{$\nu$ILG}
\newcommand{\nilgConstSize}{\scriptsize}
\newcommand{\nilgFont}[1]{\texttt{#1}}
\newcommand{\objt}{\nilgFont{objt}}
\newcommand{\object}{{\nilgConstSize\nilgFont{object}}}
\newcommand{\pred}{\nilgFont{pred}}
\newcommand{\funcc}{\nilgFont{func}}
\newcommand{\comp}{\nilgFont{comp}}
\newcommand{\achv}{\nilgFont{achv}}
\newcommand{\apg}{{\nilgConstSize\nilgFont{apg}}}
\newcommand{\upg}{{\nilgConstSize\nilgFont{upg}}}
\newcommand{\apv}{{\nilgConstSize\nilgFont{apn}}}
\newcommand{\ang}{{\nilgConstSize\nilgFont{ang}}}
\newcommand{\ung}{{\nilgConstSize\nilgFont{ung}}}

\newcommand{\wl}{\textsc{WL}}

\newcommand{\mhn}{\ensuremath{\text{M}(3h\concat3n)}}
\newcommand{\hmrp}{\ensuremath{h^{\text{mrp}}}}
\newcommand{\hff}{\ensuremath{h^{\text{FF}}}}

\title{
    WLPlan: Relational Features for Symbolic Planning
}
\author{Dillon Z. Chen}
\affiliations{LAAS-CNRS, University of Toulouse\\dchen@laas.fr}
\nocopyright

\begin{document}

\maketitle

\begin{abstract}
Scalable learning for planning research generally involves juggling between
different programming languages for handling learning and planning modules
effectively. Interpreted languages such as Python are commonly used for learning
routines due to their ease of use and the abundance of highly maintained
learning libraries they exhibit, while compiled languages such as C++ are used
for planning routines due to their optimised resource usage. Motivated by the
need for tools for developing scalable learning planners, we introduce WLPlan, a
C++ package with Python bindings which implements recent promising work for
automatically generating relational features of planning tasks. Such features
can be used for any downstream routine, such as learning domain control
knowledge or probing and understanding planning tasks. More specifically, WLPlan
provides functionality for (1) transforming planning tasks into graphs, and (2)
embedding planning graphs into feature vectors via graph kernels. The source
code and instructions for the installation and usage of WLPlan are available at
\url{tinyurl.com/42kymswc}
\end{abstract}

\section{Introduction}
Learning to plan has regained significant interest in recent years due to advancements of machine learning (ML) approaches across various fields.
An aim of learning to plan involves designing automated, domain-independent algorithms for learning domain knowledge from small training problems used for scaling up planning to problems of very large sizes~\cite{toyer.etal.2018,toyer.etal.2020,dong.etal.2019,shen.etal.2020,karia.srivastava.2021,staahlberg.etal.2022,mao.etal.2023,chen.etal.2024a}.
A recent state-of-the-art approach involves learning heuristic functions from relational features automatically extracted from graph representations of planning tasks for both classical~\cite{chen.etal.2024} and numeric~\cite{chen.thiebaux.2024} planning.
The approach involves (1) transforming planning tasks into graphs, and (2) embedding such graphs into feature vectors.
Its performance can be attributed to its evaluation speed and expressive power over previous methods.

\begin{figure}
    \centering
    \newcommand{\cheat}{-0.0cm}
    \newcommand{\segmentWidth}{1.65cm}
    \newcommand{\segmentHeight}{3.35cm}
    \newcommand{\segmentShift}{3.51cm}
    \input{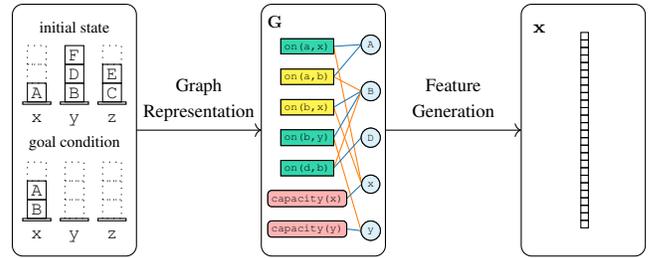}
    \vspace*{\cheat}
    \caption{The core components of the WLPlan package. (Numeric) planning tasks are transformed into graphs, which are then automatically embedded into feature vectors.}
    \vspace*{\cheat}
    \label{fig:wlplan}
\end{figure}

We introduce the WLPlan package which implements various transformations of planning tasks into graphs, and graph kernels for embedding such graphs.
Figure~\ref{fig:wlplan} illustrates these routines.
The underlying algorithms are implemented in C++ for optimised runtime and memory usage, and Python bindings are provided for easy prototyping and training of ML models.
Most learning for planning architectures are either written in Python and heavily unoptimised or conversely written entirely in C++ and difficult to extend and prototype with.
Furthermore, WLPlan can be extended to support new graph representations of planning tasks and graph kernels.
This paper serves several purposes:
\begin{enumerate}[label=(\arabic*)]
    \item 
    We provide the general algorithm, functionalities and design principles of the WLPlan package for streamlining learning for planning research.
    \item We describe currently implemented graph representations and graph kernels in WLPlan, some of which have not been published in the papers it was built upon.
    \item We outline various practical uses of WLPlan, including data visualisation, distinguishability testing, and learning heuristic functions, with experiments and analysis.
\end{enumerate}

\begin{figure*}[t!]
    \centering
    \newcommand{\moduleShift}{12cm}
    \newcommand{\xShift}{2.25}
    \newcommand{\modA}{11cm}
    \newcommand{\modB}{9cm}
    \scalebox{0.7}{
        \input{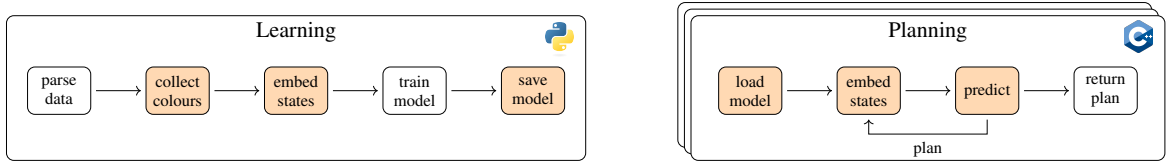}
    }
    \caption{
        A typical learning for planning pipeline and corresponding routines.
        Learning is usually done once, but planning is done many times.
        The learning module is usually implemented in Python for rapid prototyping and training with common ML libraries.
        The planning module is usually implemented in C++ for optimised planning. 
        WLPlan routines are highlighted in {\color{orange}{orange}} and can be called from both Python and C++.
        The ``embed states'' routine used in both modules is illustrated in Figure~\ref{fig:wlplan}.
    }
    \label{fig:pipeline}
\end{figure*}

The structure of the remainder of the paper is as follows.
Sec.~\ref{sec:wlplan} highlights the features and design choices of the current version of WLPlan, summarised in Figures~\ref{fig:wlplan} and \ref{fig:pipeline}.
Sec.~\ref{sec:background} provides the necessary technical background required for the remainder of the paper.
Sec.~\ref{sec:graph-representation} describes the currently implemented graph representations of planning tasks in the WLPlan package.
Sec.~\ref{sec:feature-generation} presents the currently implemented graph kernels for converting graphs into feature vectors.
Sec.~\ref{sec:experiments} illustrates various uses of WLPlan and experiments in learning for planning.
%

\section{WLPlan}
\label{sec:wlplan}

State-of-the-art ML techniques for planning primarily consist of graph learning approaches due to their ability to handle relational information in planning tasks, as well as their ability to operate on arbitrary task sizes.
Such approaches summarily follow a three step process:
\begin{enumerate}[label=(\alph*)]
    \item Convert a planning task to a graph.
    \item Pass the graph through a graph learning model.
    \item Learn some form of domain knowledge for planning.
\end{enumerate}
Despite this simple three step process, performing research in the field is arduous due to the burden of implementing both learning and planning modules.
Both modules are inherently nontrivial to handle concurrently as researchers have to balance code optimisation, extensibility, and communication across modules.
For example, Python is the go-to programming language for ML research due to its versatility and the existence of large collection of open-source ML libraries accessible via Python~\cite{raschka.etal.2020,github.2022}, but planning in Python is slow and inefficient.
Conversely, planners are implemented in C++ for optimised runtime and memory usage, but support for and implementation of ML models in C++ is limited and difficult.
Lastly, models trained in Python have to be serialised and compatible with C++ code.

WLPlan bridges this gap by providing a C++ package with Python bindings for automatically generating feature vectors of planning tasks.
Feature vectors are easy to handle across programming languages as well as learning and planning modules as they are lightweight and agnostic to any downstream task.
In contrast, deep learning models require more implementation effort across various modules and are restricted to the task they are designed for.
Recent work by~\citet{chen.etal.2024} also showed that for (b), it is sufficient to restrict our attention to classical ML approaches involving feature generators compared to deep learning approaches such as graph neural networks (GNNs), as the former are superior to the latter across various learning and planning metrics.
WLPlan exploits this observation and focuses on the construction and design of cheap ML models that can be implemented easily in both learning and planning modules.

\subsection{What WLPlan brings to the table}
We highlight the core functionalities WLPlan aims to offer for both learning and planning, complemented with example code snippets in both Python and C++.
It handles the feature generation and serialisation aspects of a typical learning for planning pipeline as illustrated in Figure~\ref{fig:pipeline}.

\sssection{Graph representations of planning tasks}
WLPlan includes implementations of graph transformations of planning tasks with a simple graph interface that can be accessed from both C++ and Python.
WLPlan graphs exhibit the base graph structure of nodes and edges, as well as node and edge features.
WLPlan can also be extended to support new graph representations of planning tasks.

\begin{lstlisting}[language=Python, 
caption={Python code for transforming a planning task (objects and goal) and state into a graph. C++ code is similar.},
label={lst:graph}
]
from wlplan.graph import ILGGenerator
  ...
ilg_generator = ILGGenerator(domain)
ilg_generator.set_problem(problem)
graph = ilg_generator.to_graph(state)
\end{lstlisting}

\sssection{Embeddings of planning tasks and graphs}
WLPlan provides a collection of graph kernels for embedding graphs into feature vectors.
Furthermore, WLPlan can also perform the steps of transforming planning tasks into graphs and embedding such graphs into feature vectors in a single step.
This allows researchers to efficiently embed of planning tasks with low amounts of code, in order to focus on other aspects of the machine learning pipeline such as data configuration, model prototyping, and parameter learning.

\begin{lstlisting}[language=Python, 
caption={Python code for collecting features and embedding a dataset of tasks or graphs. C++ code is similar.}
]
from wlplan.feature_generation import WLFeatures
  ...
model = WLFeatures(domain)
model.collect(dataset)
X = model.embed(dataset)
\end{lstlisting}

\sssection{Serialisation of models}
Lastly, WLPlan provides functionality for saving and loading models from both Python and C++.
Model information includes planning domain information that a model was designed for, the built up hash function and collected colours of graph kernels, and optionally, learned weights of linear models.
Models are both lightweight in terms of the number of parameters, and transparent as they are saved in a readable JSON format.

\begin{lstlisting}[language=Python, 
caption={Python code for saving a WLPlan model.}
]
model.set_weights(weights)
model.save(save_path)
\end{lstlisting}

\begin{lstlisting}[language=C++, 
caption={C++ code for loading a WLPlan model.}
]
#include "<path_wlplan_include>/feature_generation/feature_generator_loader.hpp"
  ...
std::shared_ptr<feature_generation::Features> model
    = load_feature_generator(save_path);
\end{lstlisting}

\subsection{What WLPlan does not aim to support}
Conversely, we also highlight the functionalities WLPlan does not aim to support and provide.

\sssection{Data extraction}
Surveys and studies~\cite{press.2016,roh.etal.2021} have shown that in the context of machine learning routines, a significant amount of time (greater than 50\%) is spent on collecting and organising datasets for training.
WLPlan only focuses on the modelling side of learning routines and thus is not concerned with data collection and organisation.
Nevertheless, WLPlan makes use of the pddl~\cite{favorito.etal.2024} package for automatically parsing planning tasks and converting them into a format fit for the WLPlan interface.

\sssection{Learning}
To keep WLPlan simple in terms of dependencies and functionality, the package does not provide any learning algorithms.
We note that vectors and graphs generated from WLPlan can be used with any downstream classical ML and deep learning model, for which there exist plenty of highly maintained, optimised and open-source libraries.

\sssection{Planning}
Similarly, WLPlan also does not provide the functionalities required for a complete planner, such as a successor generator or search algorithms.
Various planning libraries and systems already exist, with which WLPlan can be integrated for a complete planning system.

Indeed, we refer the reader to the GOOSE\footnote{Available at \url{tinyurl.com/mrxc8wjf}} planner for a complete learning for planning system which combines WLPlan and other learning and planning components.

\section{Background}
\label{sec:background}
This section provides the formal definitions required for understanding the algorithms and technical details of WLPlan.
For the sake of brevity, we only focus on deterministic planning representations, but note that WLPlan is state-centric, meaning that it is agnostic to the transition model, and thus can also handle probabilistic or non-deterministic actions.

\subsection{Planning Task}
Let $\range{n}$ denote the set of integers $\set{1,\ldots,n}$.
A planning task can be understood as a state transition model~\cite{geffner.bonet.2013} given by a tuple $\Pi = \gen{S, A, s_0, G}$ where $S$ is a set of states, $A$ a set of actions, $s_0 \in S$ an initial state, and $G \subseteq S$ a set of goal states.
Each action $a \in A$ is a function $a: S \rightarrow S \cup \set{\bot}$ where $a(s) = \bot$ if $a$ is not applicable in $s$, and $a(s) \in S$ is the successor state when $a$ is applied to $s$.
A solution for a planning task is a plan: a sequence of actions $\pi = a_1, \ldots, a_n$ where $s_i = a_i(s_{i-1}) \not= \bot$ for $i \in \range{n}$ and $s_n \in G$.
A state $s$ in a planning task $\Pi$ induces a new planning task $\Pi' = \gen{S, A, s, G}$.
A planning task is solvable if there exists at least one plan.

\subsection{Numeric Planning}
Numeric planning, formalised in PDDL2.1~\cite{fox.long.2003}, is a compact, lifted representation of a planning task using predicate logic and relational numeric variables.
More specifically, a numeric planning task is a tuple $\Pi = \gen{\objects, \predicates, \functions, \schemata, s_0, \goal}$, where $\objects$ denotes a set of objects, $\predicates$/$\functions$ a set of predicate/function symbols, $\schemata$ a set of action schemata, $s_0$ the initial state, and $\goal$ now the goal condition.
As mentioned above, understanding of the transition system induced by $\schemata$ is not required, so we instead focus on the representation of states and the goal condition next.
%

\sssection{States}
Each symbol $\sigma \in \predicates \cup \functions$ is associated with an arity $\arity(\sigma)\in\N \cup \set{0}$.
Predicates and functions take the form $p(x_1,\ldots,x_{\arity(p)})$ and $f(x_1,\ldots,x_{\arity(f)})$ respectively, where the $x_i$s denote their arguments.
Propositional and numeric variables are defined by substituting objects into predicate and function variables.
More specifically, given $\s \in \predicates \cup \functions$, and a tuple of objects $\objs = \gen{\objs_1,\ldots,\objs_{\arity(\s)}}$, we denote $\s(\objs)$ as variable defined by substituting $\objs$ into arguments of $\s$.
A state $s$ is an assignment of values in $\set{\top, \bot}$ (resp. $\R$) to all possible propositional (resp. numeric) variables in a state.
Following the closed world assumption, we can equivalently represent a state as a set of true propositions and numeric assignments.
Let $X_p(s)$ denote the set of propositional variables that are true in $s$, $X_n(s)$ the set of numeric variables, and $X(s) = X_p(s) \cup X_n(s)$.

\sssection{Goal Condition}
A propositional condition is a positive (resp. negative) literal $x=\top$ (resp. $x=\bot$) where $x$ is a propositional variable.
A numeric condition has the form $\xi \unrhd 0$ where $\xi$ is an arithmetic expression over numeric variables and $\unrhd \in \set{\geq, >, =}$.
We let $[x]^s$ (resp. $[\xi]^s$) denote the value of a numeric variable $x$ (resp. expression $\xi$) in a state $s$, and $V(\xi)$ for the set of numeric state variables in $\xi$.
The goal condition $\goal$ is a set of propositional and numeric conditions which we denote $\goal_p$ and $\goal_n$, respectively.
A state $s$ satisfies the goal condition $\goal$ if $s$ satisfies all its conditions.

\sssection{Domain}
A planning domain is a set of planning tasks which share the same set of predicates, functions and action schemata.
Constant objects are named objects that occur in all planning tasks in a domain.

\sssection{Example}
%
As a running example, we consider the Capacity Constrained Blocksworld (ccBlocksworld) domain by~\citet{chen.thiebaux.2024}.
ccBlocksworld is a numeric extension of Blocksworld where there are a finite number of bases on which blocks can be stacked, and each base has a capacity constraint on the number of blocks that can be stacked on it.
The objective of ccBlocksworld is to rearrange blocks from an initial to goal tower configuration.
The left image of Figure~\ref{fig:wlplan} illustrates a numeric ccBlocksworld planning task where there are 3 bases each with a capacity constraint of 3.
In the 4-operations encoding of ccBlocksworld, an optimal plan requires 16 actions, while only 10 actions if the bases have no capacity constraints.

\subsection{Graphs}\label{ssec:graph}
We denote a graph with categorical and continuous node features and edge labels by a tuple $\graph = \gen{\nodes, \edges, \featCat, \featCon, \featEdge}$.
We have that $\nodes$ is a set of nodes, $\edges \subseteq \nodes \times \nodes$ a set of edges, $\featCat:\nodes \to \nodeCat$ the categorical node features, $\featCon:\nodes \to \R$ are the continuous node features, and $\featEdge:\edges \to \edgeCat$ the edge labels.
The neighbourhood of a node $u \in \nodes$ in a graph is defined by $\neighbour(u) = \set{v \in \nodes \mid \gen{u,v} \in \edges}$.
The neighbourhood of a node $u \in \nodes$ with respect to an edge label $\iota$ is defined by $\neighbour_{\iota}(u) = \set{v \in \nodes \mid e=\gen{u,v} \in \edges \land \featEdge(e) = \iota}$.

\section{Graph Representation}
\label{sec:graph-representation}
The first component of the WLPlan package involves the transformation of planning tasks into graphs.
Graphs with edge features are viewed as `relational structures' in other communities, from which we can derive relational features with various sorts of algorithms.
In this section, we describe the numeric Instance Learning Graph (\nilg) for representing numeric planning tasks~\cite{chen.thiebaux.2024}, which generalises the Instance Learning Graph (ILG) for classical planning tasks~\cite{chen.etal.2024}.
The \nilg{} is the primary graph in the WLPlan package but the package can be extended to support arbitrary graph representations of planning tasks, such as those in the literature~\cite{toyer.etal.2018,shen.etal.2020,silver.etal.2021,staahlberg.etal.2022,chen.etal.2024a,horcik.sir.2024}.
%

The middle image in Figure~\ref{fig:wlplan} illustrates a subgraph of the \nilg{} encoding of the numeric ccBlocksworld planning task in the left image.
Nodes of the \nilg{} represent the objects (blue), goal conditions (yellow), and state information of the planning task, with colours encoding the semantics of nodes.
From the closed world assumption, only true propositional variables (green) are encoded as nodes, alongside all numeric variables (red).
Edges connect objects to variables that are predicates or functions instantiated with the object, as well as numeric goal conditions to their numeric variables.
Edge labels encode the location of predicates and functions in which objects are instantiated.
In the image, blue/orange edges connect variable nodes to the object that is instantiated in the first/second argument.

\begin{definition}
    The numeric Instance Learning Graph (\nilg) of a numeric planning task $\Pi = \gen{\objects, \predicates, \functions, \schemata, s_0, \goal}$ is a graph with categorical and continuous node features and edges labels $\graph = \gen{\nodes, \edges, \featCat, \featCon, \featEdge}$ with
    \begin{itemize}
        \item nodes $\nodes = \objects \cup X(s_0) \cup \goal$ where we assume w.l.o.g. that $V(g) \subseteq X(s_0)$ for all $g \in \goal_n$.
        \item edges $\edges = \bigcup_{p=\sigma(\textbf{o}) \in X(s_0) \cup \goal} \set{\gen{p, \mathbf{o}_i} \mid i \in \range{\arity(\sigma)}} \cup$ \newline
        \hspace*{2cm}$\bigcup_{(\xi \unrhd 0) \in \goal_n} \set{\gen{\xi, v} \mid v \in V(\xi)}$.
        \item categorical node features $\featCat: \nodes \to \nodeCat$ defined by
        \begin{align*}
            \featCat(u) \!=\!
            \begin{cases}
                \objt(u) &\text{if $u \in \objects$} \\
                (\pred(u), \apg) &\text{if $u \in X_p(s_0) \cap \goal_p$} \\
                (\pred(u), \upg) &\text{if $u \in \goal_p \setminus X_p(s_0)$} \\
                (\pred(u), \apv) &\text{if $u \in X_p(s_0) \setminus \goal_p$} \\
                \funcc(u) &\text{if $u \in X_n(s_0)$} \\
                (\comp(u), \achv(u)) &\text{if $u \in \goal_n$}
            \end{cases}
        \end{align*}
        where $\objt(u) = u$ if $u$ is a constant object and $\object$ otherwise, $\pred(u)/\funcc(u)$ denote the predicate/function symbol of a propositional/numeric variable $u$, $\comp(u) \in \set{\geq, >, =}$ encodes the comparator type of the numeric goal condition $u$, and $\achv(u) \in \set{\ung, \ang}$ encodes whether $s_0$ satisfies $u$.
        We note that $\object$, alongside $\apg, \upg, \apv, \ang, \ung$ are constant categorical node features\footnote{Standing for achieved propositional goal, unachieved propositional goal, achieved propositional nongoal, achieved numeric goal, and unachieved numeric goal, respectively.}.
        \item continuous node features $\featCon: \nodes \to \R$ defined by
        \begin{align*}
            \featCon(u) =
            \begin{cases}
                [u]^{s_0} &\text{if $u \in X_n(s_0)$} \\
                [\xi]^{s_0} &\text{if $u = (\xi \unrhd 0) \in \goal_n, [\xi]^{s_0}\!\!\not\!\unrhd 0$} \\
                0 &\text{otherwise}
            \end{cases}
        \end{align*}
        \item edge labels $\featEdge: \edges \to \N \cup \set{0}$ defined by
        \begin{align*}
            \featEdge(e) \!=\!
            \begin{cases}
                i &\text {if $e = \gen{p, \mathbf{o}_i}$ for $p=\sigma(\mathbf{o}) \in X(s_0) \cup \goal_p$} \\
                0 &\text {otherwise, i.e. $e = \gen{\xi, v}$ for $(\xi \unrhd 0) \in \goal_n$}
            \end{cases}
        \end{align*}
    \end{itemize}
\end{definition}

In general, the number of categorical node features for a domain with predicates $\predicates$ and functions $\functions$ is $\abs{\nodeCat} = 1 + \abs{\text{constant\_objects}} + 3\abs{\predicates} + \abs{\functions} + 4$.
Continuous node features indicate the value of numeric variables in the state and the error in the expression of numeric goals if it has not been achieved in $s_0$, and are set to zero for all other nodes.

\section{Feature Generation}
\label{sec:feature-generation}

\begin{figure*}
    \newcommand{\vcnt}[1]{\raisebox{-0.5\height}{#1}}
    \centering
    \vcnt{\scalebox{0.75}{
        \trimbox{0.15cm 1.8cm 15.5cm 1.8cm}{
            \begin{tikzpicture}

    \newcommand{\majorSize}{3.8cm}
    \newcommand{\majorShift}{4.5cm}
    \newcommand{\minorShift}{2.5cm}
    \newcommand{\nodeShift}{1cm}

    \newcommand{\circSize}{0.5cm}
    \newcommand{\dualSize}{1.5cm}
    \newcommand{\sqrtTwo}{1.41421356237}

    \newcommand{\minorSize}{0.5}
    \newcommand{\minorNudge}{1.12cm}
    \newcommand{\minorEdgeNudge}{0.55cm}

    \tikzset{
        badge/.style={
            rectangle,
            rounded corners,
            draw,
        },
        major/.style={
            badge,
            minimum width=\majorSize,
            minimum height=\majorSize,
        },
        node/.style={
            draw,
            circle,
            inner sep=0,
            minimum size=0.3cm,
        },
        minorNode/.style={
            node,
            minimum size=0.3cm*\minorSize,
        },
        circleNode/.style={
            badge,
            minimum width=\circSize,
            minimum height=\circSize,
        },
        minorCircleNode/.style={
            rectangle,
            draw,
            rounded corners=2pt,
            minimum size=\minorSize*\circSize,
        },
        dottedNode/.style={
            circleNode,
            dashed,
        },
        minorDottedNode/.style={
            minorCircleNode,
            dotted,
        },
        title/.style={
            align=left,
            anchor=north west,
        },
        bottom/.style={
            align=left,
            anchor=south west,
            font=\tiny,
            white,
        }
    }

    \newcommand{\tc}[1]{\tikz[baseline=-0.5ex]\draw[radius=1.5pt,fill=#1] (0,0) circle ;}%
    \renewcommand{\tc}[1]{}

    \newcommand{\lime}{1.9cm}
    \newcommand{\titleCoord}{-\lime, \lime}
    \newcommand{\bottomCoord}{-\lime, -\lime}
    \newcommand{\aCoord}{0, 0}
    \newcommand{\bCoord}{0, \nodeShift}
    \newcommand{\cCoord}{\nodeShift, 0}
    \newcommand{\dCoord}{0, -\nodeShift}
    \newcommand{\eCoord}{-\nodeShift, 0}

    \newcommand{\gcol}{green!20}
    \newcommand{\bcol}{blue!20}
    \newcommand{\ocol}{orange!20}
    \newcommand{\pcol}{purple!20}

    \newcommand{\graphMacro}{
        \node[node,fill=\gcol]  (a) at (\aCoord) {};
        \node[node,fill=\gcol]  (b) at (\bCoord) {};
        \node[node,fill=\bcol]   (c) at (\cCoord) {};
        \node[node,fill=\ocol] (d) at (\dCoord) {};
        \node[node,fill=\pcol] (e) at (\eCoord) {};

        \draw(a) -- (b);
        \draw(a) -- (c);
        \draw(a) -- (d);
        \draw(c) -- (d);
        \draw(b) -- (e);
    }

    \begin{scope}
        \node[major] at (0, 0) {};
        \node[title] at (\titleCoord) {WL};
        \node[bottom] at (\bottomCoord) {$(\tc{\gcol}, \mseta{\tc{\gcol}, \tc{\ocol}, \tc{\bcol}})$};
        \graphMacro{}
        \node[circleNode] at (\aCoord) {};
        \node[dottedNode] at (\bCoord) {};
        \node[dottedNode] at (\cCoord) {};
        \node[dottedNode] at (\dCoord) {};
    \end{scope}

    \begin{scope}[xshift=1*\majorShift]
        \renewcommand{\mseta}[1]{\{ \!\! \{ \! #1 \! \} \!\! \}}
        \node[major] at (0, 0) {};
        \node[title] at (\titleCoord) {2-WL};
        \graphMacro{}
        \node[circleNode, minimum width=\dualSize] at ($(a)!0.5!(c)$) {};
        \node[dottedNode] at (\bCoord) {};
        \node[dottedNode] at (\dCoord) {};
        \node[dottedNode] at (\eCoord) {};
        \node[bottom] at (\bottomCoord) {$(\!\mseta{\tc{\gcol}, \tc{\pcol}},\!\mseta{
        \mseta{\tc{\gcol},\!\tc{\bcol},\!\tc{\gcol}}
        ,\!\mseta{\tc{\gcol},\!\tc{\bcol},\!\tc{\ocol}}
        ,\!\mseta{\tc{\gcol},\!\tc{\bcol},\!\tc{\pcol}}
        }\!)$};
    \end{scope}

    \begin{scope}[xshift=2*\majorShift]
        \node[major] at (0, 0) {};
        \node[title] at (\titleCoord) {2-LWL};
        \graphMacro{}
        \node[circleNode, minimum width=\dualSize] at ($(a)!0.5!(c)$) {};
        \node[dottedNode] at (\bCoord) {};
        \node[dottedNode] at (\dCoord) {};
        \node[bottom] at (\bottomCoord) {$(\!\mseta{\tc{\gcol}, \tc{\pcol}},\!\mseta{
            \mseta{\tc{\gcol},\!\tc{\bcol},\!\tc{\gcol}}
            ,\!\mseta{\tc{\gcol},\!\tc{\bcol},\!\tc{\ocol}}
        }\!)$};
    \end{scope}

    \begin{scope}[xshift=3*\majorShift]
        \node[major] at (0, 0) {};
        \node[title] at (\titleCoord) {iWL};

        \node[minorCircleNode] (center) at (\aCoord) {};
        \coordinate[left=\minorNudge of center] (lcenter) ;
        \coordinate[right=\minorNudge of center] (rcenter) ;
        \coordinate[above=\minorNudge of center] (acenter) ;
        \coordinate[below=\minorNudge of center] (bcenter) ;
        \node[minorCircleNode] at (lcenter) {};
        \node[minorCircleNode] at (rcenter) {};
        \node[minorCircleNode] at (acenter) {};
        \node[minorCircleNode] at (bcenter) {};

        \renewcommand{\nodeShift}{0.5cm}
        \node[minorDottedNode] (center) at (\bCoord) {};
        \coordinate[left=\minorNudge of center] (lcenter) ;
        \coordinate[right=\minorNudge of center] (rcenter) ;
        \coordinate[above=\minorNudge of center] (acenter) ;
        \coordinate[below=\minorNudge of center] (bcenter) ;
        \node[minorDottedNode] at (lcenter) {};
        \node[minorDottedNode] at (rcenter) {};
        \node[minorDottedNode] at (acenter) {};
        \node[minorDottedNode] at (bcenter) {};
        \node[minorDottedNode] (center) at (\cCoord) {};
        \coordinate[left=\minorNudge of center] (lcenter) ;
        \coordinate[right=\minorNudge of center] (rcenter) ;
        \coordinate[above=\minorNudge of center] (acenter) ;
        \coordinate[below=\minorNudge of center] (bcenter) ;
        \node[minorDottedNode] at (lcenter) {};
        \node[minorDottedNode] at (rcenter) {};
        \node[minorDottedNode] at (acenter) {};
        \node[minorDottedNode] at (bcenter) {};
        \node[minorDottedNode] (center) at (\dCoord) {};
        \coordinate[left=\minorNudge of center] (lcenter) ;
        \coordinate[right=\minorNudge of center] (rcenter) ;
        \coordinate[above=\minorNudge of center] (acenter) ;
        \coordinate[below=\minorNudge of center] (bcenter) ;
        \node[minorDottedNode] at (lcenter) {};
        \node[minorDottedNode] at (rcenter) {};
        \node[minorDottedNode] at (acenter) {};
        \node[minorDottedNode] at (bcenter) {};

        \renewcommand{\nodeShift}{1cm}
        \begin{scope}[xshift=3*\majorShift]
            \scalebox{\minorSize}{\graphMacro{}}
            \begin{scope}[xshift=\minorShift]
                \scalebox{\minorSize}{\graphMacro{}}
            \end{scope}
            \begin{scope}[xshift=-\minorShift]
                \scalebox{\minorSize}{\graphMacro{}}
            \end{scope}
            \begin{scope}[yshift=\minorShift]
                \scalebox{\minorSize}{\graphMacro{}}
            \end{scope}
            \begin{scope}[yshift=-\minorShift]
                \scalebox{\minorSize}{\graphMacro{}}
            \end{scope}
        \end{scope}

        \node[minorNode, fill=white, draw=white] (center) at (\aCoord) {};
        \coordinate[left=\minorNudge+\minorEdgeNudge of center] (lcenter) ;
        \coordinate[right=\minorNudge+\minorEdgeNudge of center] (rcenter) ;
        \coordinate[above=\minorNudge+\minorEdgeNudge of center] (acenter) ;
        \coordinate[below=\minorNudge+\minorEdgeNudge of center] (bcenter) ;
        \node[minorNode, fill=white, draw=white] at (lcenter) {};
        \node[minorNode, fill=white, draw=white] at (rcenter) {};
        \node[minorNode, fill=white, draw=white] at (acenter) {};
        \node[minorNode, fill=white, draw=white] at (bcenter) {};
        \node[font=\tiny] at (center) {$\otimes$};
        \node[font=\tiny] at (lcenter) {$\otimes$};
        \node[font=\tiny] at (rcenter) {$\otimes$};
        \node[font=\tiny] at (acenter) {$\otimes$};
        \node[font=\tiny] at (bcenter) {$\otimes$};

    \end{scope}

\end{tikzpicture}
        }
    }}
    \quad
    \vcnt{\scalebox{0.8}{\begin{tikzpicture}
    \newcommand{\size}{1.25cm}
    \tikzset{
        group/.style={
                rectangle,
                draw,
                rounded corners,
                minimum width=\size,
                minimum height=\size,
            },
    }
    \node[anchor=south,group] at (0, 0) {WL};
    \node[anchor=south,group, minimum width=1.2*\size, minimum height=1.6*\size, text depth=1.4cm] at (0, -0.1*\size) {2-LWL};
    \node[anchor=south,group, minimum width=1.3*\size, minimum height=2.1*\size, text depth=2.05cm] at (0, -0.15*\size) {2-WL};
    \node[anchor=west, group, minimum height=1.1*\size, minimum width=2.2*\size, align=right, text width=23mm] at (-0.5*\size-0.05*\size, 0.5*\size) {ccWL};
    \node[anchor=east, group, minimum height=1.1*\size, minimum width=2.2*\size, align=left, text width=23mm] at (0.5*\size+0.05*\size, 0.5*\size) {iWL};
\end{tikzpicture}}}
    \caption{
        Left: Visualisations of neighbours (dotted) of the center node or node pair (solid) in the WL, 2-WL, 2-LWL and iWL algorithms.
        %
        %
        In iWL, the WL algorithm is run $\abs{\nodes}$ times, where each time a different node is individualised with a special colour.
        Right: Expressivity hierarchy of WL algorithms implemented in WLPlan.
    }
    \label{fig:wls}
    \label{fig:hierarchy}
\end{figure*}
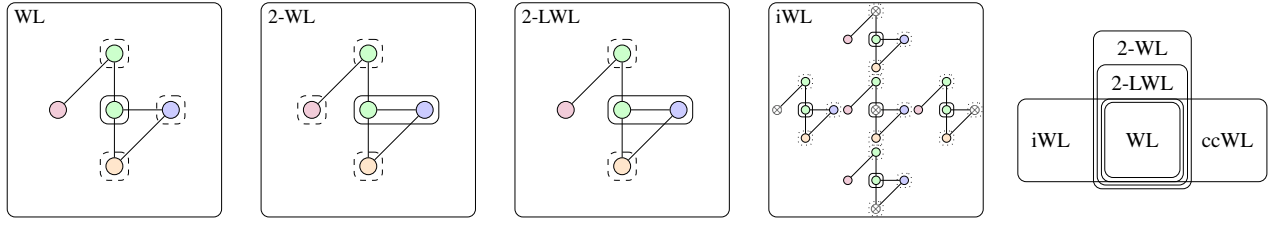

The second component of the WLPlan package involves transforming graph representations of planning tasks into feature vectors for use with any downstream task.
The go-to example is learning heuristic functions, although it is also possible to perform all sorts of other tasks as we will discuss later in Section~\ref{sec:experiments}.

The algorithms for feature generation of graphs are generally some extension of the colour refinement algorithm, a special case of the general $k$-Weisfeiler-Leman algorithm~\cite{weisfeiler.leman.1968,cai.etal.1989}.
In this section, we begin with describing the colour refinement, or 1-Weisfeiler-Leman (WL) algorithm, followed by how the algorithm is used for constructing feature vectors of graphs.
Lastly, we outline extensions of WL that been have implemented thus far in WLPlan.

\subsection{The WL Algorithm}
The underlying concept of the WL algorithm is to iteratively update node colours based on the colours of their neighbours.
The original WL algorithm was designed for graphs without edge labels.
We present the WL algorithm which can support edge labels~\cite{barcelo.etal.2022} in Algorithm~\ref{alg:wl}.
The algorithm's input is a graph with node features and edge labels as described in Section~\ref{ssec:graph}, alongside a hyperparameter $L$ determining how many WL iterations to perform.
The output of the algorithm is a canonical form for the graph that is invariant to node orderings.

Line 1 initialises node graph colours as their categorical node features.
Lines 2 and 3 iteratively update the colour of each node $v$ in the graph by collecting all its neighbours and the corresponding edge label $(u, \iota)$ into a multiset.
This multiset is then hashed alongside $v$'s current colour with an injective function to produce a new refined colour.
In practice, the injective hash function is built lazily, where every time a new multiset is encountered, it is mapped to a new, unseen hash value.
After $L$ iterations, the multiset of all node colours seen throughout the algorithm is returned.

\newcommand{\algoFontSize}{\small}
\begin{algorithm}
    \algoFontSize
    \caption{WL algorithm}\label{alg:wl}  
\KwData{A graph $\graph = \gen{\nodes, \edges, \featCat, \featCon, \featEdge}$, injective $\hash$ function, and number of iterations $L$.}  
\KwResult{Multiset of colours.}  
$c^{0}(v) \la \featCat(v), \forall v \in \nodes$ \label{line:wl:init}\\
\For{$j=1,\ldots,L$ \normalfont{\textbf{do for}} $v \in \nodes$}{ 
    $c^{j}(v) \la 
    \hash
    \lr{c^{j-1}(v), 
    \bigcup_{\iota\in\edgeCat}\mseta{(c^{j-1}(u), \iota) \mid u \in \neighbour_{\iota}(v)}}
    $ \label{line:wl:update}
} 
\Return{$\bigcup_{j=0,\ldots,L}\mseta{c^{j}(v) \mid v \in \nodes}$} \label{line:wl:return}

\end{algorithm}
\begin{algorithm}
    \algoFontSize
    \caption{2-WL algorithm}\label{alg:2wl}  
\KwData{A graph $\graph = \gen{\nodes, \edges, \featCat, \featCon, \featEdge}$, injective $\hash$ function, and number of iterations $L$.}  
\KwResult{Multiset of colours.}  
$e(v, u) \la \featEdge(v, u), \forall (v,u) \in \edges$ \\
$e(v, u) \la \bot, \forall (v, u) \in (\nodes^2) \setminus \edges$ \\
$c^{0}(v, u) \la (\featCat(v), \featCat(u), e(v, u)), \forall (v,u) \in \nodes^2$ \label{line:2wl:init}\\
\For{$j=1,\ldots,L$ \normalfont{\textbf{do for}} $(v,u) \in \nodes^2$}{ 
    $c^{j}(v,u) \la 
    \hash
    \bigl(c^{j-1}(v,u), 
    \{\!\{
    (c^{j-1}(w,u), c^{j-1}(v,w))
    \mid 
    w \in \nodes
    \}\!\}
    \bigr)
    $ \label{line:2wl:update}
} 
\Return{$\bigcup_{j=0,\ldots,L}\mseta{c^{j}(v, u) \mid (v, u) \in \nodes^2}$} \label{line:2wl:return}

\end{algorithm}
\begin{algorithm}
    \algoFontSize
    \newcommand{\nodePair}[1]{\seta{#1}}
\caption{2-LWL algorithm}\label{alg:2lwl}  
\KwData{
    An undirected graph $\graph = \gen{\nodes, \edges, \featCat, \featCon, \featEdge}$, injective $\hash$ function, and number of iterations $L$.
    Let $\nodePair{u, v}$ denote a pair of nodes without order or undirected edge.
}  
\KwResult{Multiset of colours.}  
$e\nodePair{v, u} \la \featEdge\nodePair{v, u}, \forall \nodePair{v, u} \in \edges$ \\
$e\nodePair{v, u} \la \bot, \forall \nodePair{v, u} \in {{\nodes}\choose{2}} \setminus \edges$ \\
$c^{0}\nodePair{v, u} \!\la\! (\featCat(v), \featCat(u), e\nodePair{v, u}), 
\forall \nodePair{v, u} \!\in\! {{\nodes}\choose{2}}$ 
\label{line:2lwl:init}\\
\For{$j=1,\ldots,L$ \normalfont{\textbf{do for}} $\nodePair{v, u} \in {{\nodes}\choose{2}}$}{ 
    $c^{j}\nodePair{v, u} \la 
    \hash
    \bigl(c^{j-1}\nodePair{v, u}, 
    \{\!\{
    \mseta{c^{j-1}\nodePair{w, u}, c^{j-1}\nodePair{v, w}}
    \mid 
    w \in \neighbour(v) \cup \neighbour(u)
    \}\!\}
    \bigr)
    $ \label{line:2lwl:update}
} 
\Return{${\displaystyle}_{j=0,\ldots,L}\mseta{c^{j}\nodePair{v, u} \mid \nodePair{v, u} \in {{\nodes}\choose{2}}}$} \label{line:2lwl:return}

\end{algorithm}
\begin{algorithm}
    \algoFontSize
    \caption{iWL algorithm}\label{alg:iwl}  
\KwData{A graph $\graph = \gen{\nodes, \edges, \featCat, \featCon, \featEdge}$, injective $\hash$ function, and number of iterations $L$.}  
\KwResult{Multiset of colours.}  
\For{$w \in \nodes$}{
    $c^{0}_w(v) \la \featCat(v), \forall v \in \nodes \setminus \set{w}$ \label{line:iwl:init}\\
    $c^{0}_w(w) \la \otimes$  \\
    \For{$j=1,\ldots,L$ \normalfont{\textbf{do for}} $v \in \nodes$}{ 
        $c^{j}_w(v) \la 
        \hash
        \lr{c^{j-1}_w(v), 
        \bigcup_{\iota\in\edgeCat}\mseta{(c^{j-1}_w(u), \iota) \mid u \in \neighbour_{\iota}(v)}}
        $ \label{line:iwl:update}
    } 
}
\Return{$\bigcup_{w \in \nodes}\bigcup_{j=0,\ldots,L}\mseta{c^{j}_w(v) \mid v \in \nodes}$} \label{line:iwl:return}

\end{algorithm}
\begin{algorithm}
    \algoFontSize
    \caption{ccWL algorithm}\label{alg:ccwl}  
\KwData{A graph $\graph = \gen{\nodes, \edges, \featCat, \featCon, \featEdge}$, injective $\hash$ function, number of iterations $L$, and aggregator $\oplus$.}  
\KwResult{Multiset of colours, and function $\featCol$ mapping colours to vectors in $\R^d$.}  
$c^{0}(v) \la \featCat(v), \forall v \in \nodes$ \label{line:ccwl:init}\\
\For{$j=1,\ldots,L$ \normalfont{\textbf{do for}} $v \in \nodes$}{ 
    $c^{j}(v) \la 
    \hash
    \lr{c^{j-1}(v), 
    \bigcup_{\iota\in\edgeCat}\mseta{(c^{j-1}(u), \iota) \mid u \in \neighbour_{\iota}(v)}}
    $ \label{line:ccwl:update}
}
$\Mst \la \bigcup_{j=0,\ldots,L}\mseta{c^{j}(v) \mid v \in \nodes}$ \\
$\nodes(c) \!\la\! \set{v \!\in\! \nodes \!\mid\! \exists j \in \set{0}\!\cup\!\range{L}, c^j(v) = c}\!, \forall c \!\in\! \Mst$ \\
$\featCol(c) \la \bigoplus_{v \in \nodes(c)} \featCon(v), \forall c \in \Mst$ \label{line:ccwl:featcol}\\
\Return{$(\Mst, \featCol)$} \label{line:ccwl:return}

\end{algorithm}

\subsection{Feature Vectors from WL}
The WL algorithm has been used to generate features for the WL graph kernel~\cite{shervashidze.etal.2011}.
Each node colour constitutes a feature, and the value of a feature for a graph is the count of the number of nodes that exhibit or has exhibited the colour.
Then given a set of colours $\colours$ known a priori, the WL algorithm can return a fixed sized feature vector of size $\abs{\colours}$ for every input graph.
We first describe how to collect the colours $\colours$ from a set of given planning tasks, followed by how to embed arbitrary graphs into fixed sized feature vectors from such colours.

\sssection{Collecting colours}
We can construct $\colours$ from a given set of graph representations of planning tasks $\graph_1, \ldots, \graph_m$ by running the WL algorithm, with the same $\hash$ function and number of iterations $L$, on all of them and then taking the set union of all multiset outputs, i.e. $\colours = \bigcup_{i\in\range{m}} \wl(\graph_i)$.

\sssection{Embedding graphs}
Now suppose we have collected a set of colours and enumerated them by $\colours = \seta{c_1, \ldots, c_{\abs{\colours}}}$.
Then given a graph $\graph$ and its multiset output from the WL algorithm $\Mst$, we can define an embedding of the graph into Euclidean space by the feature vector
\begin{align}
    [\countCol(\Mst, c_1), \ldots, \countCol(\Mst, c_{\abs{\colours}})] \in \R^{\abs{\colours}},
    \label{eq:embed}
\end{align}
where $\countCol(\Mst, c_i)$ is an integer which counts the occurrence of the colour $c_i$ in $\Mst$.
We note importantly that any colours returned in $\Mst$ that are not in $\colours$ are defined as \emph{unseen} colours and are entirely ignored in the output.

\subsection{WL Extensions}
The WL algorithm is the canonical graph kernel baseline for graph learning tasks due to the theoretical result that it upper bounds distinguishing power of the message passing GNN architecture~\cite{morris.etal.2019,xu.etal.2019}.
It is also an efficient algorithm that runs in low polynomial time in the input graph and considered the first choice to apply to graphs as described in the extensive graph kernel survey by~\citet{kriege.etal.2020}.
Nevertheless, the graph learning community has proposed various extensions of the WL algorithm and corresponding GNN architectures that have provably more distinguishing power than the WL algorithm, and yet are still computationally feasible.

Most notably, it is well known that the $(k+1)$-WL algorithm is strictly more powerful than the $k$-WL algorithm for $k \geq 2$ but its runtime scales exponentially in $k$.
Thus, many extensions of the WL algorithm and corresponding GNNs have been proposed to either approximate or achieve orthogonal expressiveness of higher order WL algorithms~\cite{morris.etal.2020,morris.etal.2022,zhao.etal.2022a,wang.etal.2023,alvarezgonzalez.etal.2024}.
Furthermore, graph kernels have been proposed that also handle graphs with continuous node attributes~\cite{chen.thiebaux.2024}.
We have implemented some of these WL extensions alongside completely new graph kernels in the current version of WLPlan which we outline as follows.
Figure~\ref{fig:hierarchy} illustrates the expressivity hierarchy of mentioned WL algorithms.


\sssection{2-WL}
The $k$-WL algorithms are a suite of incomplete graph isomorphism algorithms which have a one-to-one correspondence to $k$-variable counting logics~\cite{cai.etal.1989}.
However, the $k$-WL algorithms scale exponentially in $k$, with the 2-WL algorithm exhausting memory limits on medium sized graph datasets.
We describe and implement the 2-WL algorithm and refer to~\cite[Section 5]{cai.etal.1989} for the general $k$-WL algorithm.

The idea of the 2-WL algorithm, outlined in Algorithm~\ref{alg:2wl}, is to now assign and refine colours to ordered pairs of nodes.
The algorithm begins in Lines 1-3 by assigning all possible ordered pairs of nodes a tuple of the node colours as well as the edge label between them.
If there is no edge between a pair of nodes, a special $\bot$ value is used as the edge label.
Lines 4-5 then iteratively refine the colour of each node pair by defining the neighbour of a pair $(v, u)$ to be a \emph{set of tuple of pairs} $((w, u), (v, w))$ where $w$ ranges over all nodes in the graph.
Then the algorithm applies the colouring function of the current iteration to all node pairs to create a multiset of tuples of colours which are then hashed alongside the current node pair's colour.
Finally, the algorithm returns the multiset of all node pair colours seen throughout the algorithm in Line 6.
To generalise to the $k$-WL algorithm, one replaces node pairs with node $k$-tuples.

\sssection{2-LWL}
The $k$-LWL algorithms~\cite{morris.etal.2017} provide efficient approximations of the $k$-WL algorithms but still have the same worst case computational complexity.
The main approximations made are that node tuples are converted to node sets, reducing the number of possible node tuples to consider per iteration by a constant factor ($n^k \to {{n}\choose{k}}$), and relaxing the definition of neighbours of node tuples.
Now the neighbour for a 2-sets of nodes $\set{v, u}$ in 2-LWL is defined by the set of set of 2-sets $\set{\set{w, u}, \set{v, w}}$ where $w$ now ranges over the union of neighbours of $u$ and $v$, instead of over all nodes.
Algorithm~\ref{alg:2lwl} outlines the 2-LWL algorithm and Figure~\ref{fig:wls} illustrates the different neighbour definitions of 2-WL and 2-LWL.

\sssection{iWL}
We introduce an expressive WL algorithm extension inspired by Identity-aware GNNs (ID-GNNs)~\cite{you.etal.2021}, orthogonal to the $k$-WL algorithms.
ID-GNNs run a GNN $\abs{\nodes}$ times on a graph which uses different parameters for embedding updates on a selected, individualised node on each GNN run.
We kernelise the ID-GNN algorithm into what we call the individualised WL algorithm, presented in Algorithm~\ref{alg:iwl} and Figure~\ref{fig:wls}.

We have a single outer loop iterating over all nodes $w \in \nodes$ in a graph in Line 1, and within each inner loop, all nodes are assigned their initial node colour, except for $w$ which is assigned a special, individualised colour $\otimes$ that is not in $\nodeCat$ of $\featCat$.
The remainder of the algorithm is equivalent to the WL algorithm, except that iWL now returns $\abs{\nodes}$ more colours in the output multiset due to the outer loop.


\sssection{ccWL}
Lastly, we outline a WL algorithm which can handle both categorical and continuous node features, as all WL algorithms covered thus far only handle categorical node features.
This is the ccWL algorithm used for numeric planning introduced by~\citet{chen.thiebaux.2024}.
Although there exist other WL algorithms that have been proposed to handle continuous node features~\cite{morris.etal.2016,togninalli.etal.2019}, they lose the semantics of numbers required for reasoning in numeric planning.
Also differently to the aforementioned WL algorithms, the ccWL algorithm returns both a multiset of colours and a function $\featCol$ which maps colours to continuous features.
The collection of output continuous features can then be concatenated to the embeddings of multisets described in Equation~\eqref{eq:embed}.

The algorithm is outlined in Algorithm~\ref{alg:ccwl} and is equivalent to WL for Lines 1-4.
Line 5 then assigns each seen colour $c$ the set of nodes $\nodes(c)$ in the graph that exhibited the colour through the WL algorithm.
Line 6 then aggregates the continuous features of nodes in $\nodes(c)$ for each colour with a given function $\oplus$ to produce the continuous feature $\featCol(c)$.

\section{Example Uses}
\label{sec:experiments}

In this section, we outline various ways of using WLPlan with corresponding experiments and results.

\subsection{Data Visualisation}
Several theoretical frameworks have been developed to better understand the behaviour of planning domains such as novelty width~\cite{lipovetzky.geffner.2012}, correlation complexity~\cite{seipp.etal.2016}, the river measure~\cite{dold.helmert.2024}, and methods to bound such measures~\cite{dold.helmert.2024a}.
Automatic tools include~\citeauthor{hoffmann.2011}'s~(\citeyear{hoffmann.2011}) Torchlight for testing the difficulty of planning domains based on the existence of local minima in the search space with respect to $h^+$~\cite{hoffmann.2005}, and hand-crafted planning features for constructing planning portfolios~\cite{ferber.seipp.2022} that can be analysed to understand what features of a planning task are well suited for a specific planner. 
In light of tools for probing and understanding planning domains without any knowledge of them a priori, we propose using Principal Component Analysis (PCA) visualisations of WLPlan embeddings to understand certain structures of planning tasks.

\subsubsection{Setup}
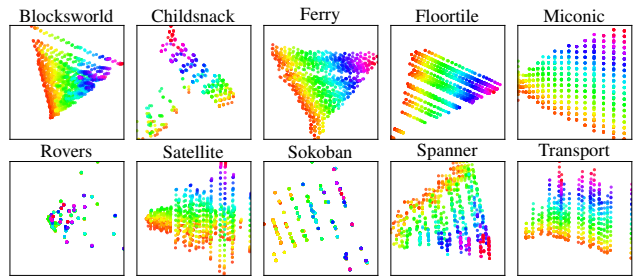
\begin{figure}
    \centering
    \offinterlineskip
    \newcommand{\pcaWidth}{0.09\textwidth}
    \newcommand{\insertFig}[2]{\begin{tikzpicture}[every node/.style={inner sep=0,outer sep=0}]
            \draw (0, 0) node {\includegraphics[width=\pcaWidth]{figures/pca/#1.pdf}};
            \draw (0, 0.9) node {\scriptsize #2};
        \end{tikzpicture}%
}
    \insertFig{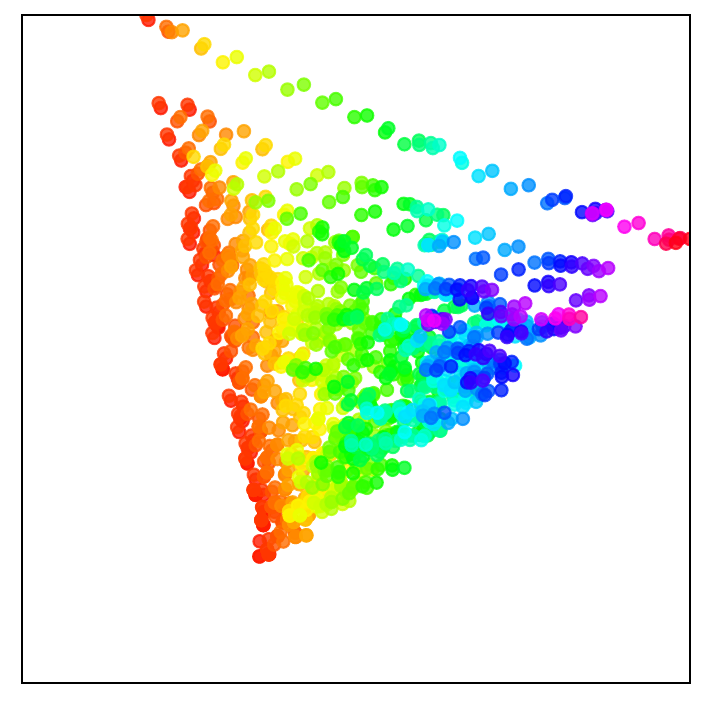}{Blocksworld}
    \insertFig{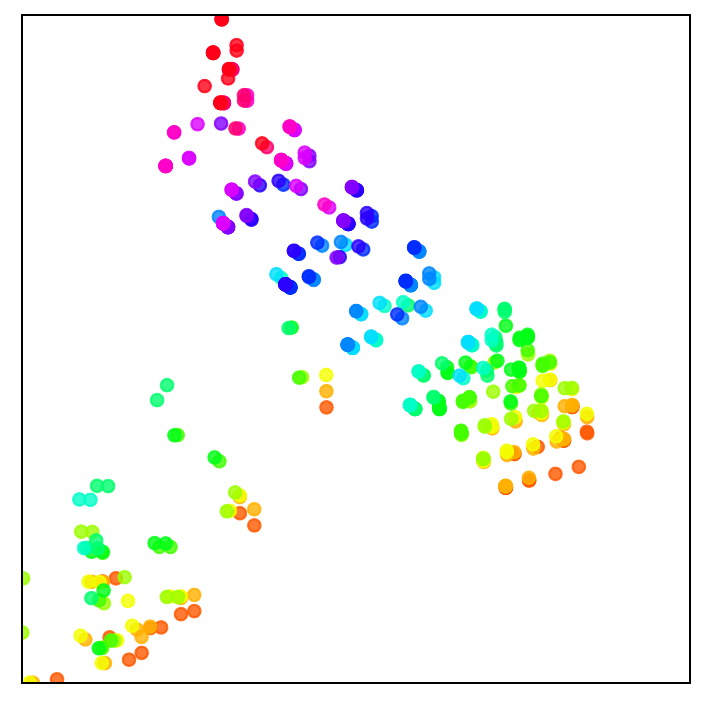}{Childsnack}
    \insertFig{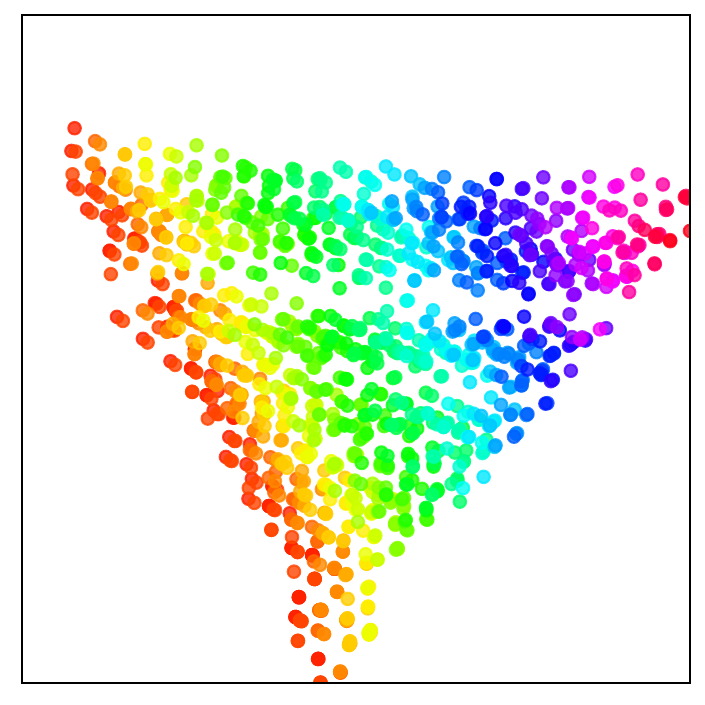}{Ferry}
    \insertFig{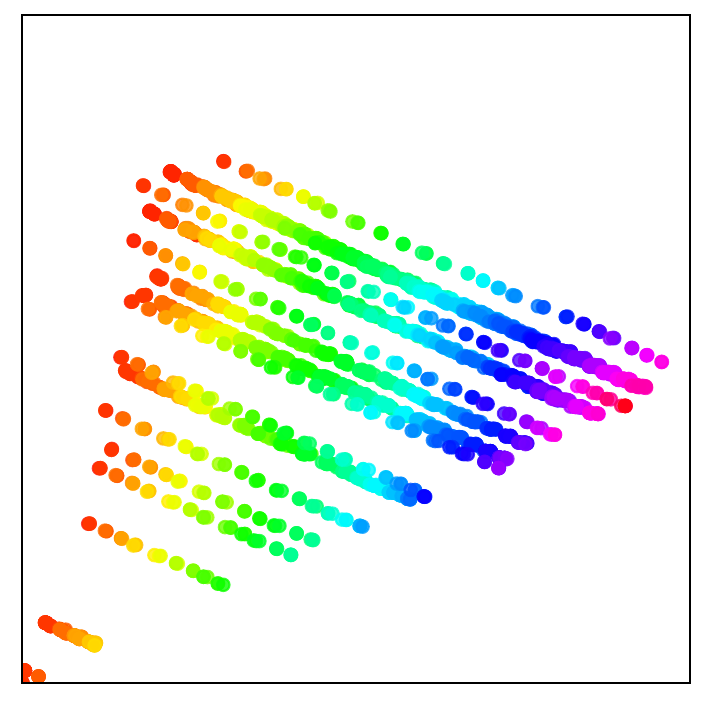}{Floortile}
    \insertFig{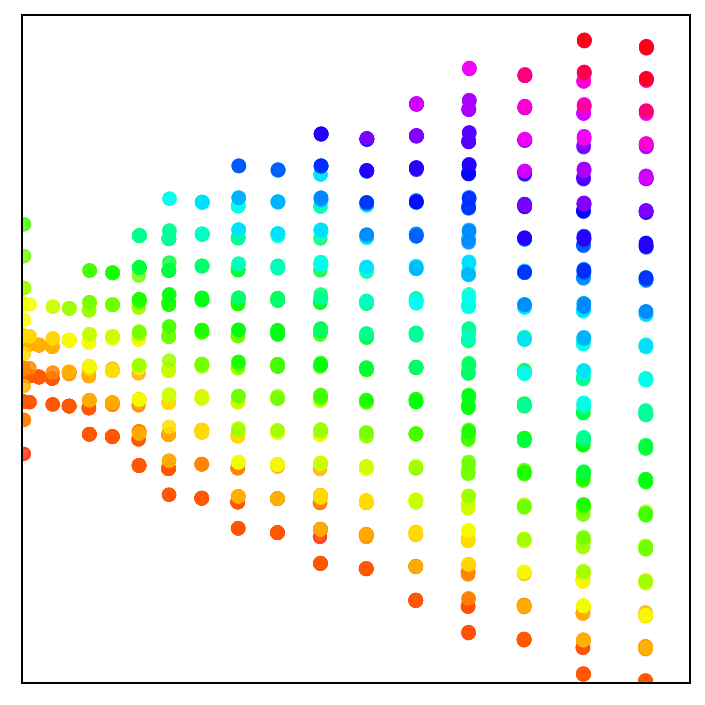}{Miconic}
    \insertFig{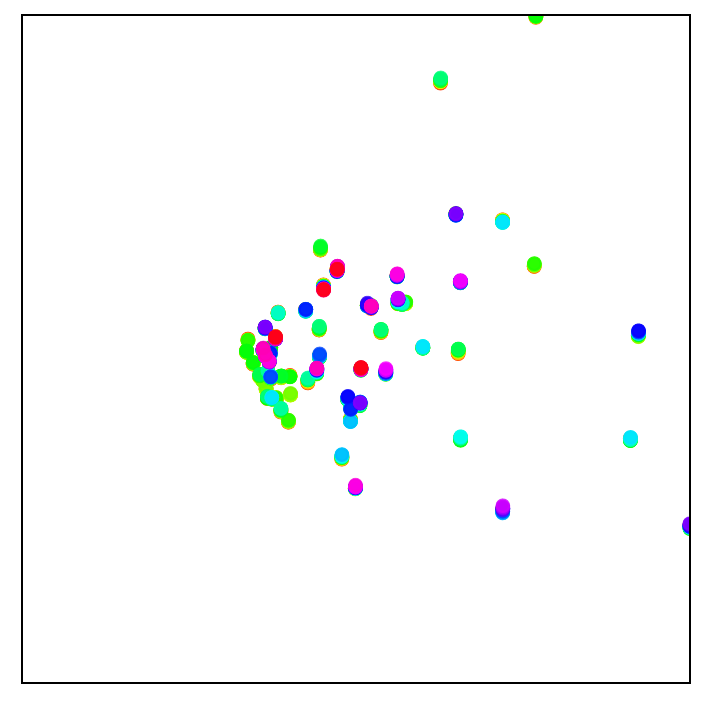}{Rovers}
    \insertFig{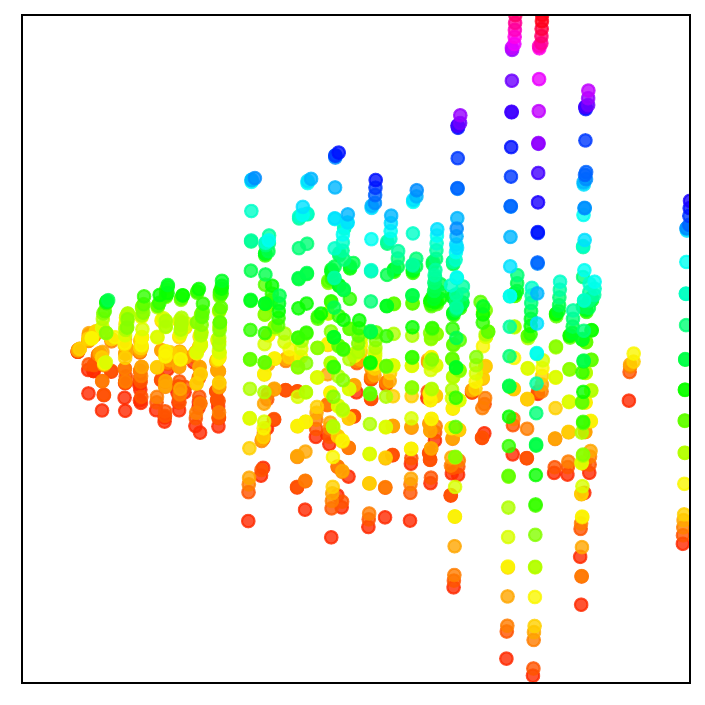}{Satellite}
    \insertFig{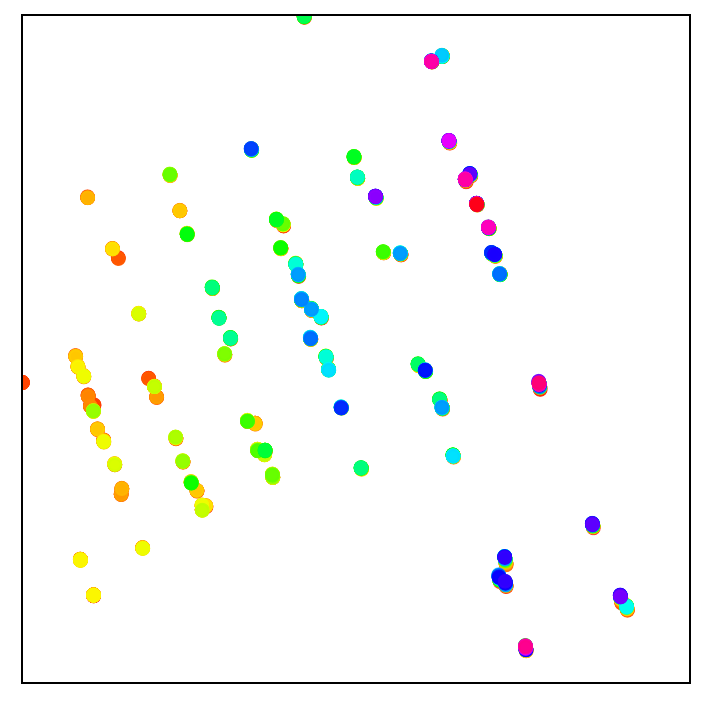}{Sokoban}
    \insertFig{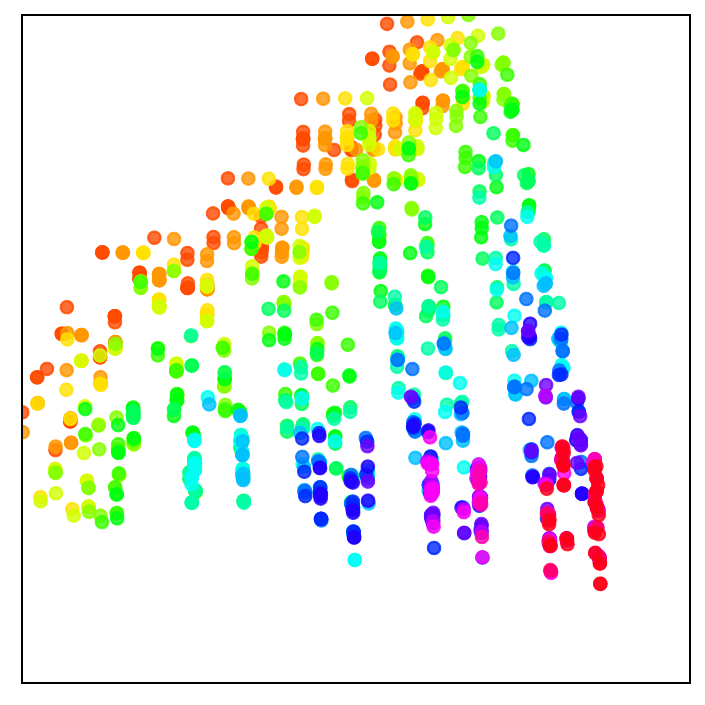}{Spanner}
    \insertFig{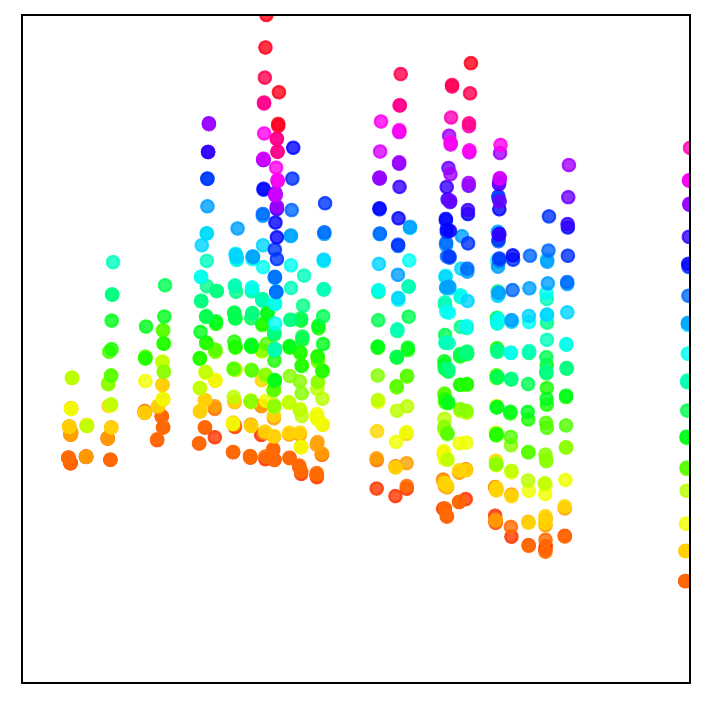}{Transport}
    \caption{
        PCA visualisation of WL embeddings of training states for IPC23LT domains
        Red/purple on one end of the rainbow scale indicates lower/higher $h^*$ values.
    }
    \label{fig:visualisation}
\end{figure}

PCA is a dimensionality reduction technique that projects high-dimensional data into lower-dimensional space through maximising the variance of projected features, see~\cite[Sec. 12.1]{bishop.2007} for more details.
Thus, we can use PCA to visualise WL embeddings of planning states and the distribution of $h^*$ values.
Given a set of $n$ planning states from optimal training plan traces of the Learning Track of the 2023 International Planning Competition (IPC23LT)~\cite{taitler.etal.2024}, we collect all colours and embed the states into Euclidean space.
The embedded states can be represented by a matrix $\X \in \R^{n \times d}$ with rows corresponding to states and columns WL features.
We can then run PCA on $\X$ to get a projected matrix $\X' \in \R^{n \times 2}$, where columns correspond to the two principal components.
Figure~\ref{fig:visualisation} illustrates $\X'$ for each domain in the IPC23LT, colouring each point by the state's $h^*$ value.

\subsubsection{Results}
Visualising PCA projections of WL embeddings can give us an idea on the relationship between planning features and $h^*$ values, as well as the informativeness of the training set.
Indeed, we see that some domains \emph{appear} to convey a linear relationship with the $h^*$ values such as in Blocksworld, Ferry, and Miconic, which correlates with the high performance of learned linear heuristics on these domains~\cite{chen.etal.2024}.
Conversely, we note that projected embeddings on some domains exhibit no clear pattern such as in Rovers and Sokoban which suggests that the features may not be informative enough to learn a good heuristic function.
The sparsity of points in these domains as well as in Childsnack may also suggest the need for more training data.
For example the training set for Childsnack consists of at most one child allergic to gluten, whereas this number if arbitrary in the testing set.

\subsection{Distinguishability Tests}
In line with tools for probing and understanding planning tasks, we make use of distinguishability tests that are commonly performed in ML literature~\cite{kriege.etal.2020,abboud.etal.2021,balcilar.etal.2021,feng.etal.2022,zhao.etal.2022,wang.etal.2023,bouritsas.etal.2023,horcik.sir.2024,drexler.etal.2024a} for testing whether GNN architectures and WL algorithms are expressive enough to distinguish various graph representations of planning tasks with different $h^*$ values.
The WLPlan package allows for quick implementations of such setups by making use of the theoretical fact that WL algorithms subsume GNN architectures for graph isomorphism testing~\cite{morris.etal.2019,xu.etal.2019}.

\subsubsection{Setup}
To perform a distinguishability test, we collect all colours in a given set of $n$ planning states from optimal plan traces of training tasks for each IPC23LT domain, with which we use to embed all such states into Euclidean space.
Next, we check how many of the ${n \choose 2}$ pairs of vectors are distinguished by observing whether their outputs are different.
The metric used to evaluate the tests is to count how many pairs of graphs are not distinguished by a model, with 0 being the best achievable score.
We perform experiments with the WL, 2-LWL and iWL algorithm and $L=2$.
We do not display results for 2-WL as it uses significantly more memory compared to other considered WL algorithms.

\subsubsection{Results}
From the results in Table~\ref{tab:distinguishability}, we note that the WL algorithm is able to distinguish all pairs of training states for 7 out of 10 domains.
The 2-LWL algorithm ran out of memory for 4 out of 10 domains while collecting colours, and the iWL algorithm on 1 domain.
Despite using significantly more computation, we note that the more expressive WL algorithms are able to distinguish more pairs of training states on one out of 10 domains they were able to terminate on.


\newcommand{\tablefontsize}{}
\begin{table}
    \centering
    \newcommand{\oom}{--}
    \tablefontsize
    \begin{tabularx}{\linewidth}{l*{10}{Y}}
\toprule
& \header{Blocksworld} & \header{Childsnack} & \header{Ferry} & \header{Floortile} & \header{Miconic} & \header{Rovers} & \header{Satellite} & \header{Sokoban} & \header{Spanner} & \header{Transport} \\
\midrule
WL & 0 & 0 & 0 & 22 & 0 & 0 & 0 & 129 & 68 & 0 \\ 2-LWL & 0 & 0 & 0 & \oom & 0 & \oom & \oom & \oom & 14 & 0 \\ iWL & 0 & 0 & 0 & 22 & 0 & \oom & 0 & 110 & 68 & 0 \\ 
\bottomrule
\end{tabularx}
    \caption{
        Number of pairs of indistinguishable states encoded as ILGs with respect to $h^*$ values with WL algorithms on IPC23LT domains.
        \oom{} indicates that the memory limit was reached while embedding the entire dataset with the corresponding WL algorithm.
        Lower ($\downarrow$) values are better.
    }
    \label{tab:distinguishability}
\end{table}

\subsection{Learning Heuristic Functions}
The most practical usage of learning for planning is to actually use learning to plan.
Our last set of experiments involve reimplementations and reproducibility checking of work by the author with WLPlan.
Specifically, we use WLPlan to reimplement the learning and usage of heuristic functions for classical planning in~\cite{chen.etal.2024} and the submission to NeurIPS-24 before acceptance for numeric planning~\cite{chen.thiebaux.2024}.

\subsubsection{Setup}
For the interest of space, we refer to the original papers for full setup details of the experiments, and only summarily highlight the results of the original papers and of the WLPlan reimplementations here.
Table~\ref{tab:classic-coverage} presents results for both classical and numeric encodings of planning domains from the IPC23LT.
WL$_{\text{ICAPS-24}}$ and GNN$_{\text{ICAPS-24}}$ refer to the original implementations of learned WL and GNN heuristics with GBFS and WL$_{\text{WLPlan}}$ refers to the WLPlan reimplementation.
Planner baselines include the $\hff$ heuristic with GBFS, and LAMA.

Numeric encodings of IPC23LT domains are either equivalent to the original domain's semantics, or are made more difficult with additional numeric constraints.
Floortile and Sokoban were omitted as there was no obvious benefit or extension to the domain that can be derived with a numeric encoding.
ccWL$_{\text{NeurIPS-24-$\alpha$}}$ refers to the implementation submitted to NeurIPS-24 before acceptance, and ccWL$_{\text{WLPlan}}$ the WLPlan reimplementation.
Planner baselines shown are $\hmrp$ with GBFS~\cite{scala.etal.2020} and the state-of-the-art numeric planner $\mhn$~\cite{chen.thiebaux.2024a}.

\subsubsection{Results}
We observe that our reimplementation of previous approaches with the WLPlan package achieves similar results to the original approaches, with minor coverage differences due to training and search variance.

\begin{table}
    \centering
    \tablefontsize
    \newcommand{\pp}{\phantom{0}}
\begin{tabularx}{\linewidth}{l*{10}{Y}r}
\toprule
& \header{Blocksworld} & \header{Childsnack$^*$} & \header{Ferry} & \header{Floortile} & \header{Miconic} & \header{Rovers} & \header{Satellite} & \header{Sokoban} & \header{Spanner$^*$} & \header{Transport$^*$} & \header{Sum} \\
\midrule
Classic \\
\midrule
WL$_{\text{WLPlan}}$    & 74 & 40 & 77 &\pp2 & 90 & 37 & 48 & 37 & 73 & 28 & 506 \\ 
WL$_{\text{ICAPS-24}}$  & 75 & 29 & 76 &\pp2 & 90 & 37 & 53 & 38 & 73 & 29 & 502 \\
GNN$_{\text{ICAPS-24}}$ & 63 & 23 & 70 &\pp0 & 89 & 26 & 31 & 33 & 46 & 32 & 413 \\
$\hff$                  & 28 & 26 & 68 &  12 & 90 & 34 & 65 & 36 & 30 & 41 & 430 \\
LAMA                    & 61 & 35 & 68 &  11 & 90 & 67 & 89 & 40 & 30 & 66 & 557 \\
\midrule
Numeric \\
\midrule
ccWL$_{\text{WLPlan}}$          & 19 & 90 & 71 & na & 90 & 23 & 16 & na & 90 & 46 & 445 \\ 
ccWL$_{\text{NeurIPS-24-$\alpha$}}$  & 26 & 90 & 69 & na & 90 & 19 & 22 & na & 90 & 53 & 459 \\
$\hmrp$                         & 19 & 25 & 60 & na & 64 & 18 & 21 & na & 42 & 32 & 281 \\
$\mhn$                          & 23 & 53 & 57 & na & 61 & 30 & 29 & na & 76 & 40 & 369 \\
\bottomrule
\end{tabularx}

    \caption{
        Coverage of various planners and learning planners on classical (top) and numeric (bottom) encodings of IPC23LT domains. 
        We refer to the text for descriptions of planners.
        Numeric encodings that are semantically equivalent to their classical counterparts are marked $*$.
        Higher ($\uparrow$) values are better.
    }
    \label{tab:classic-coverage}
    \label{tab:numeric-coverage}
\end{table}



\section{Conclusion}
WLPlan is a new tool for learning for planning, motivated by and built from state-of-the-art approaches in the field.
The package is focused on efficiency as well as ease of use and extensibility by supporting both Python and C++.
A goal of WLPlan is to streamline research in the field by providing implementations of commonly used classical and graph learning routines for planning, and keeping up-to-date with future research built on top of it.

\section*{Acknowledgements}
The author would like to acknowledge Ryan Xiao Wang for some insights into the training dataset of the Childsnack domain used in the experiments of the paper.

\small
\bibliography{icaps25}

\end{document}